\documentclass[lettersize,journal]{IEEEtran}
\usepackage[pdftex,dvipsnames]{xcolor}  
\usepackage{pgfplots}
\usepackage{tikz}
\pgfplotsset{compat=newest}
\usepackage[colorinlistoftodos,prependcaption,textsize=tiny]{todonotes}

\IEEEoverridecommandlockouts                              

\renewcommand{\baselinestretch}{0.96}

\usepackage{amsthm}
\usepackage{graphics} 
\usepackage{epsfig} 
\usepackage{times} 
\usepackage{amsmath} 
\usepackage{amssymb}  

\usepackage[normalem]{ulem}

\usepackage[maxbibnames=99,giveninits=true]{biblatex}
\usepackage{bm}
\usepackage{bbm}
\usepackage{xcolor}
\usepackage{graphicx}
\usepackage{algorithm}
\usepackage[noend]{algpseudocode}
\usepackage{booktabs}
\usepackage{siunitx}
\usepackage{hyperref}

\usepackage{caption}
\usepackage{lipsum}
\usepackage{multirow}
\usepackage{diagbox} 
\usepackage{subcaption}

\newcommand{\R}{\mathbb{R}}

\newtheorem{definition}{\bf Definition}

\newtheorem{theorem}{\bf Theorem}

\newtheorem{remark}{\bf Remark}

\begin{document}
\title{\LARGE \bf
 Efficient Greedy Algorithms for Feature Selection in Robot Visual Localization}
\author{Vivek Pandey, Amirhossein Mollaei, and Nader Motee
\thanks{V. Pandey, A. Mollaei and N. Motee are with the Department of Mechanical Engineering and Mechanics, Lehigh University. {\tt\small \{vkp219,ammb23,motee\}@lehigh.edu}.\endgraf
}
}

\maketitle

\thispagestyle{plain}
\pagestyle{plain}

\begin{abstract}

Robot localization is a fundamental component of autonomous navigation in unknown environments. Among various sensing modalities, visual input from cameras plays a central role, enabling robots to estimate their position by tracking point features across image frames. However, image frames often contain a large number of features, many of which are redundant or uninformative for localization. Processing all features can introduce significant computational latency and inefficiency. This motivates the need for intelligent feature selection—identifying a subset of features that are most informative for localization over a prediction horizon. In this work, we propose two fast and memory-efficient feature selection algorithms that enable robots to actively evaluate the utility of visual features in real time. Unlike existing approaches with high computational and memory demands, the proposed methods are explicitly designed to reduce both time and memory complexity while achieving a favorable trade-off between computational efficiency and localization accuracy. 
\end{abstract}





\section{Introduction}\label{sec:Intro}
Navigation is a fundamental aspect of robotics, essential for enabling robots to autonomously perform a wide range of tasks. Effective navigation relies on continuous and accurate position estimation, commonly referred to as localization. To achieve reliable and precise localization, robots employ advanced probabilistic estimation techniques such as the Kalman filter~\cite{thrun2005_probabilistic_robotics}, particle filter~\cite{stachniss2014particle_filters_robotics,thrun2002particle_filter_robotics}, and information filter~\cite{thrun2003info_filter_multi_robot_slam}. These algorithms integrate sensor measurements with predictive motion models to iteratively refine and update the robot’s position estimates.

In the context of perception-based navigation, robots employ sensors such as cameras and LiDAR for localization. The large volume of features captured by these sensors makes the estimation process computationally demanding. Consequently, it is crucial to efficiently select a smaller subset of the most informative features to optimize localization performance. Extensive research has been devoted to this problem, focusing on developing efficient algorithms for perception-based feature selection~\cite{hossein2020feature_fast, jiao2022lidar_slam_feature_selection, guadagnino2022fast_sparse_lidar, lerner2007landmark_selection_task}.  

In~\cite{carlone2019attention}, the authors use convex relaxation and a greedy strategy to select the most informative features. Zhao \textit{et al.}~\cite{zhao2020good_feature_matching} propose a lazier-than-lazy greedy algorithm to address the feature selection problem by maximizing the log-determinant as an information measure for optimizing Visual SLAM performance. Zhang \textit{et al.}~\cite{zhang2015good_features_cvpr} approach Visual SLAM from a system-theoretic perspective, defining observability scores to quantify feature informativeness and employing a greedy algorithm for feature selection.

In~\cite{pandey2024randomized}, the authors propose an efficient randomized algorithm for multi-agent localization that offers probabilistic performance guarantees. It is well established that subset selection is a combinatorial optimization problem. To address this challenge, several algorithmic frameworks have been explored, including convex relaxation~\cite{joshi2009sensor_convex} and greedy methods~\cite{nemhauser1978maximizing_submodular_Set_functions}. In this work, we focus on greedy algorithms, which provide practical and computationally efficient approaches for obtaining near-optimal solutions to submodular maximization problems.

\subsection{Greedy Algorithms for Submodular Maximization}

Greedy algorithms have long been recognized for their effectiveness in addressing combinatorial optimization problems, particularly those involving submodular maximization. The seminal work of Nemhauser \textit{et al.}~\cite{nemhauser1978maximizing_submodular_Set_functions} established fundamental performance guarantees for greedy algorithms applied to monotone submodular functions, demonstrating that such algorithms can achieve a \(\left(1 - 1/e\right)\)-approximation of the optimal solution.  

Greedy algorithms have since been widely adopted across diverse domains, including sensor scheduling~\cite{shamaiah2010greedy_sensor_selection}, feature selection~\cite{carlone2019attention}, and anchor selection in SLAM~\cite{chen2022anchor}, among others. Recent advancements have focused on improving the computational and memory efficiency of greedy methods, thereby enhancing both their theoretical guarantees and practical applicability in resource-constrained settings~\cite{mirzasoleiman2015stochastic_greedy,badanidiyuru2014submodular_streaming,kazemi2019submodular_streaming}.  

Building on these developments, we leverage modern variants of greedy algorithms to address the feature selection problem in robot visual localization, aiming to balance computational efficiency with localization accuracy.

\subsection{Our Contribution}

In this paper, we address the feature selection problem under computational and memory constraints commonly encountered in robotic systems. Our main contributions are summarized as follows:

\begin{itemize}
    \item \textbf{Stochastic Greedy Feature Selection:} We propose a stochastic greedy algorithm that significantly reduces the time complexity of feature selection while maintaining near-optimal performance. This approach enables efficient operation in real-time scenarios without substantial loss in accuracy.
    
    \item \textbf{Approximate Greedy Feature Selection:} We develop a mathematically justified approximate greedy algorithm that simultaneously improves both computational and memory efficiency. This formulation leverages tractable surrogates for information-based metrics to further accelerate the selection process while preserving theoretical soundness.
    
\end{itemize}
Our contributions are theoretical and analytical in nature, aimed at improving memory efficiency and fast computation.
 Some of the results in this paper have been developed independently in \cite{vafaee2025nonsubmodularvisualattentionrobot}.


\subsection{Mathematical Notation}

Vectors and matrices are denoted by lowercase and uppercase letters, respectively. Sequences of vectors or matrices stacked over time are represented by boldface lowercase and uppercase letters. The cones of symmetric positive semidefinite and positive definite \(n \times n\) matrices are denoted by \(S^n_{+}\) and \(S^n_{++}\). For a matrix \(X\), \(X^\top\) denotes its transpose. The special orthogonal group in \(\mathbb{R}^3\) is \(\mathrm{SO}(3)\), the cardinality of a set \(A\) is \(|A|\), and \(\mathrm{Tr}(\cdot)\) denotes the trace. \(I_{(\cdot)}\) represents the identity matrix, with size inferred from context.  
For \(X_1, X_2 \in S^n_{++}\), \(X_2 \preceq X_1\) if \(X_1 - X_2 \in S^n_{+}\). A Gaussian random vector \(z\) with mean \(\mu\) and covariance \(\Sigma\) is \(z \sim \mathcal{N}(\mu, \Sigma)\). The sets of nonnegative integers and reals are denoted by \(\mathbb{Z}_+\) and \(\mathbb{R}_+\), respectively.

\section{Problem Statement}\label{sec:problems_tatement}
 We address the problem of sparse feature selection to estimate the positions of a robot navigating in an unknown environment. Using efficient greedy algorithms, we select a set of the most informative features by anticipating their importance over a fixed time horizon. 

Let $\bm{x}_{t} \in \R^{3}$ be the position vector of robot at time $t \in \mathbb{Z}_+$. Consider the vector of discrete time horizon given by $[t:t+M] = [t, t+1, \cdots, t+M]$. Then vector
$\textbf{x}_{t: t+M}$
which contains states of the robot for all $t \in [t:t+M]$ can be written as 
\begin{equation}\label{eqn: robot_state_vectors_stacked}
    \textbf{x}_{t: t+M} = \left[x_t^T, x_{t+1}^T, \cdots, x_{t+M}^T\right]^T \in \R^{3N\left(M+1\right)}.  
\end{equation}

Let $\Theta_t$ represent the set of all features $f$ available at a given time step \(t\). Furthermore, let the matrix 
$\textbf{H}_{t: t + M}^f$
be the contribution of new features to the information matrix, then the information matrix is updated as follows
\begin{equation}\label{eqn:info_vec_mat_update_def}
\begin{aligned}
    \textbf{H}_{t: t + M}(\Theta_t) &= \Bar{\textbf{H}}_{t: t + M} + \sum_{f \in \Theta_t}^{}\textbf{H}_{t: t + M}^f.
\end{aligned}
\end{equation}


Given the constraints on onboard computational resources, it is essential for robot  to select a smaller subset of features \(\Phi_t\)  from the available set \(\Theta_t,\) which prioritizes the most informative features, maximizing the contribution to the information gain while minimizing the computational complexity. The feature selection problem can be formulated as the following optimization problem:
\begin{equation}\label{eqn:subset_selection}
    \Phi_t^{*} = \underset{\Phi_t \subseteq \Theta_t : |\Phi_t|\leq q}{\textup{argmax}}\rho\left(\Phi_t\right),
\end{equation}
where the goal is to maximize a non-negative scalar function \(\rho\) over the subsets of size \(q.\)

The rest of the paper is organized as follows. Section~\ref{sec:models} presents the models for robot motion and the vision system. We formulate the feature selection problem as optimal experiment design problem in Section \ref{sec:toed}. In Section~\ref{sec:greedy_algorithm}, we formally introduce our algorithms, providing a detailed description along with theoretical guarantees and analysis. 
Section~\ref{sec:conclusion} concludes the paper with a discussion of future research directions. The proofs of all theoretical results are provided in the Appendix.

\section{Model for Robot Motion and Vision System}\label{sec:models}

To achieve accurate position estimates for the robot, the robot continuously updates the initial estimates obtained from the robot dynamics using information gathered from sensor measurements. This section begins by introducing the robot motion model, which mathematically describes how the robot positions evolve over time. This model lays the foundation for incorporating measurement models, which will be used to refine the position estimates based on sensor data.

\subsection{Model for Robots Motion}

We consider that  
 dynamics of the  robot's motion is governed by the following linear dynamical model:
\begin{equation}\label{eqn:robot_dynamics_lin_ith}
    x_{\tau} = A x_{\tau-1} +  B u_{\tau} + \delta_{\tau},
\end{equation}
for all $\tau \in [t:t+M]$, where $x_{\tau}$ is the state (position vector) of the robot at time $\tau$, $ u_{\tau}$ is the control input of the  robot at time $\tau$ and $\delta_{\tau} \sim \mathcal{N}(0, \Lambda_{\tau})$ such that $\mathbb{E}[\delta_{\tau_1} \delta_{\tau_2}] = 0$ for all $\tau_1 \neq \tau_2.$

Let $\Bar{\bm{\mu}}_t  = \mathbb{E}[\bm{x}_{\tau}]$ and $\Bar{\bm{\Sigma}}_t = \mathbb{E}[\left(\bm{x}_{\tau}-\Bar{\bm{\mu}}_t\right) \left(\bm{x}_{\tau}-\Bar{\bm{\mu}}_t\right) ^T]$, then the mean and covariance of $\textbf{x}_{t:t+M}$ can be expressed as 
\begin{equation}    \label{eqn:team_mu_mat_mu_tT}
    \Bar{\bm{\mu}}_{t: t + M} =     \begin{bmatrix}
        \Bar{\bm{\mu}}_{t}& \Bar{\bm{\mu}}_{t + 1} & \cdots & \Bar{\bm{\mu}}_{t + M}\\ 
    \end{bmatrix}
\end{equation}
\begin{equation}    \label{eqn:team_cov_mat_Sigma_tT}
\Bar{\bm{\Sigma}}_{t: t + M} = 
    \begin{bmatrix}
        \Bar{\bm{\Sigma}}_{t} & \Bar{\bm{\Sigma}}_{t,t+1} & \cdots & \Bar{\bm{\Sigma}}_{t,t+M} \\
        \Bar{\bm{\Sigma}}_{t,t+1}^T & \Bar{\bm{\Sigma}}_{t+1} & \cdots & \Bar{\bm{\Sigma}}_{t+1,t+M} \\
        \vdots & \vdots & \ddots & \vdots \\
        \Bar{\bm{\Sigma}}_{t,t+M}^T & \Bar{\bm{\Sigma}}_{t+1,t+M}^T & \cdots & \Bar{\bm{\Sigma}}_{t+M}  
    \end{bmatrix}
\end{equation}
where $\Bar{\bm{\Sigma}}_{\tau} = \bm{A}\Bar{\bm{\Sigma}}_{\tau - 1} \bm{A}^T + I_N \otimes \Lambda_{\tau}$ for all $\tau \in [t: t+M]$ and \[\Bar{\bm{\Sigma}}_{\tau_1, \tau_2} = \bm{A}^{(\tau_2 - \tau_1) }  \Bar{\bm{\Sigma}}_{\tau_1}\]
for all $\tau_1, \tau_2 \in [t: t+M]$ with $\tau_1 < \tau_2.$

The information matrix of $\textbf{x}_{t:t+M}$ using the dynamical model can be written as 
\begin{equation}\label{eqn:team_initial_info_mat}
    \Bar{\textbf{H}}_{t:t+M} = \Bar{\bm{\Sigma}}_{t: t + M}^{-1}.
\end{equation}
The information matrix $\Bar{\textbf{H}}_{t:t+M}$ is a measure of information content of $\textbf{x}_{t:t+M}$ when the system dynamics is predicted over the fixed time horizon $M.$ 
The following subsections delve into the process of sensor fusion. Here, we explore how information matrices, derived from the robot motion model, are dynamically updated by incorporating measurements from the robot's sensors.

\subsection{Model for Vision System}

Let us denote the set of all features available at any time step by $\Theta_t$,  the rotation matrices describing the orientation of the camera and robot by $R_c \in $ SO(3)  and $R_{\tau} \in$ SO(3) respectively, the position vector of the feature $f \in \Theta_t$ by $y_f \in \R^3$, the position vector of the camera with respect to robot by $x_c$, and the unit vector (with respect to camera) corresponding to pixel measurement of feature $f \in \Theta_t$ at time $\tau$ by $(u_{\tau}^{f})^T \in \R^3.$
For a given vector $(u_{\tau}^{f})^T $, its cross product with another vector $v$ can be written as the product of a skew-symmetric matrix $ U_{\tau}^{f}$ and vector $v$ as 
\begin{align*}
    (u_{\tau}^{f})^T \times v =  U_{\tau}^{f} v.
\end{align*}
Then the noisy vision model for the robot proposed in \cite{carlone2019attention} is given by 
\begin{equation}\label{eqn: noisy_vision_model_ith}
    U_{\tau}^{f}(R_{\tau} R_c)^T \left(y_f - ({x}_{\tau} + R_{\tau} x_c)  \right) = {\eta}_{\tau}^{f}. 
\end{equation}
The model in \eqref{eqn: noisy_vision_model_ith} can be written equivalently as
\begin{equation}\label{eqn: noisy_vision_model_ith_alt}
    {z}_{\tau}^{f} = U_{\tau}^{f}(R_{\tau}^i R_c)^T  \left({x}_{\tau} - y_f \right) + {\eta}_{\tau}^{f}, 
\end{equation}
where ${\eta}_{\tau,T}^{f}$ is the measurement noise such that ${\eta}_{\tau,T}^{f} \sim \mathcal{N}(0,\sigma^2 I_3)$ and $ {z}_{\tau}^{f} = (U_{\tau}^{f})^T (R_c)^Tx_c.$

The vision model for the team of robots can be written by stacking the model \eqref{eqn: noisy_vision_model_ith_alt} for all robots as 
\begin{equation}\label{eqn: vision_model_team}
    {z}_{\tau}^f = F_{\tau}^f\textbf{x}_{t : t + M}+ E_{\tau}^f y_f  + {\eta}_{\tau}^f, 
\end{equation}
for appropriate matrices $F_{\tau}^f$ and $E_{\tau}^f.$

We assume that robots are capable of running $M-$step forward simulations to determine the number of frames $n_f$ in which the feature $f \in \Theta_t$ is visible.
The overall $M-$step vision model for a given feature $f$ can be written by vertically stacking \eqref{eqn: vision_model_team} as 
\begin{equation}\label{eqn: camera_model_forward}
    \textbf{z}_{t: t + M}^f = \textbf{F}_{t : t + M}^{f} \textbf{x}_{t : t + M} + \textbf{E}_{t: t + M}^{f} y_f + \bm{\eta}_{t: t + M}^f.
\end{equation}
Let us denote the covariance matrix of $\bm{\eta}_{t:t+M}^f$ by 
\begin{equation}\label{eqn:vision_noise_covaraince}
    \mathbb{E}\left[\bm{\eta}_{t: t+M}^f \left(\bm{\eta}_{t:t+M}^{f}\right)^T \right] = \sigma^2 I_{3n_f}\in S^{3n_f}_{+}.
\end{equation}
The information matrix of the parameters $\textbf{x}_{t : t + M}$ and $y_f$ is given by 
\begin{equation} \label{eqn: Omega_mat_1}
    \Omega_{\star}^{f} = \sigma^{-2}
            \begin{bmatrix}
         \left(\textbf{F}_{\star}^{f}\right)^T\textbf{F}_{\star}^{f} & \left(\textbf{F}_{\star}^{f}\right)^T\textbf{E}_{\star}^{f}\\
         \left(\textbf{E}_{\star}^{f}\right)^T \textbf{F}_{\star}^{f} & \left(\textbf{E}_{\star}^{f}\right)^T\textbf{E}_{\star}^{f}
    \end{bmatrix},
\end{equation}
where $\star = t:t+M$ is introduced for simplicity of notation.
The information matrix corresponding to the parameters $\textbf{x}_{t : t + M}$ is obtained by taking the Schur complement of the block $\left(\textbf{E}_{\star}^{f}\right)^T\textbf{E}_{\star}^{f}$ and can be written as 
\begin{equation}\label{eqn:Hf_mat}
    \scalebox{0.85}{$
        \hspace{-0.3cm}\textup{\bf{H}}_{\star}^f = \sigma_i^{-2}\Bigg(\left(\textbf{F}_{\star}^{f}\right)^T\textbf{F}_{\star}^{f} - \left(\textbf{F}_{\star}^{f}\right)^T\textbf{E}_{\star}^{f}\left(\left(\textbf{E}_{\star}^{f}\right)^T \textbf{E}_{\star}^{f}\right)^{-1}\left(\textbf{E}_{\star}^{f}\right)^T\textbf{F}_{\star}^{f}\Bigg).
    $}
\end{equation}
The information matrix $\textup{\bf{H}}_{\star}^f = \textup{\bf{H}}_{t:t+M}^f$ specified in \eqref{eqn:Hf_mat} is a measure of amount of information contained by a feature $f$ for estimation of $\textbf{x}_{t : t + M}.$

\section{Theory of Optimal Experiment Design}\label{sec:toed}
We forumlate the feature selection problem \eqref{eqn:subset_selection} for robot's localization as parameter estimation problem given a set of observations. Specifically, for estimating robot's position given \(n\) visual features, we choose the features that decreases the covraince of robot's position the most. Using the Cramer-Rao bound \cite{rao1945information,cramer1946mathematical}
\begin{equation}\label{eqn:cramer_rao_bound}
    \text{var}\left(\hat{\theta}\right) \succeq {I(\theta)^{-1}},
\end{equation}
which states that variance in the eastimated parameter \(\hat{\theta}\) is lower bounded by information matrix \(I(\theta)\). Equivalently, to minimize a scalar measure of covaraince matrix, theory of optimal design of experiments suggests maximixing the scalar measure of information matrix \cite{pukelsheim2006optimal}. 
For any new observations by tracking a feature $f$, the information matrix of $\textbf{x}_{t:t+M}$, is updated as 
\begin{equation} \label{eqn:info_matrix_update_feature_f}
    \textbf{H}_{\star} \left(\{f\}\right) = \Tilde{\textbf{H}}_{\star} + \textup{\bf{H}}_{\star}^f.
\end{equation}
Since each feature's contribution is independent, the updates to the corresponding information matrices for a set $\Phi_t$ of selected features can be done by adding the individual information matrices according to the following:
\begin{equation}\label{eqn:infor_mat_update_feature}
    \textbf{H}_{\star} \left(\Phi_t\right) = \Tilde{\textbf{H}}_{\star} + \sum_{f \in \Phi_t}^{}\textup{\bf{H}}_{\star}^f.
\end{equation}
After accounting for all visible features in the set $\Phi_t,$ the information matrix can be inverted to obtain the covariance matrix according to
\begin{equation}\label{eqn:convert_inform_mat_to_covar}
    \bm{\Sigma}_{\star}(\Phi_t) =  \textbf{H}_{\star} \left(\Phi_t\right)^{-1}.
\end{equation}
To select a feature, it must be triangulated and the information matrix $\left(\textbf{E}_{\star}^{f}\right)^T\textbf{E}_{\star}^{f}$ corresponding to its position vector $y_f$ must be invertible.

To identify the most informative features, we define suitable metrics that quantify information content or estimated accuracy. To evaluate the information content of features, several performance measures can be employed. We discuss some of these measures next.
\begin{table}[t]
    
    \centering
    
        \caption{Performance Measures}
        {\renewcommand{\arraystretch}{2}
    \begin{tabular}{lcc}
    \hline
    \hline
        Performance Measures & Matrix Operator Form \\
        \hline
        Variance of the Error $\rho_v(\textup{\textbf{H}}_{\star}(\Phi_t))$ & $\textup{Tr}\left(\textup{\textbf{H}}_{\star}\left(\Phi_t\right)^{-1}\right)$\\
        Differential Entropy $\rho_e(\textup{\textbf{H}}_{\star}(\Phi_t))$ &  $- \log\left(\textup{det}\left (\textup{\textbf{H}}_{\star}(\Phi_t)\right)\right )$\\
        Spectral Variance $\rho_{\lambda}(\textup{\textbf{H}}_{\star}(\Phi_t)) $ & $\lambda_{\textup{min}}\left(\textup{\textbf{H}}_{\star}(\Phi_t)^{-1}\right)$\\
        \hline
        \hline
    \end{tabular}}
    \label{tab: Performance Measures}
\end{table}

\subsection{Performance Measures}
We quantify the informativeness of a feature subset,  $\Phi_t \subset \Theta_t$, by employing established performance measures  \cite{hossein2020feature_fast, arash2022space_time_sampling, carlone2019attention}. These performance measures are monotonic map of the covariance matrix, which is defined in \eqref{eqn:convert_inform_mat_to_covar}. For accurate robot team localization, lower values of these performance measures indicate a better estimate of the robots' positions. Specific examples of these performance measures are listed in Table \ref{tab: Performance Measures}.


\section{Efficient Greedy Algorithms}\label{sec:greedy_algorithm}
In this section, we present the details of the feature selection algorithms designed to offer an efficient and approximate solution to maximization problem \eqref{eqn:subset_selection}. To this end, we first introduce a few key definitions that will be used throughout the discussion.

\begin{definition}\label{def:submodular_set_functions}
\cite{nemhauser1978maximizing_submodular_Set_functions} A set function \(g:2^V \rightarrow \R,\) where \(V\) is a ground set, is submodular if, for any two sets \(A\subseteq B \subseteq V\) and any element \(e \in V\setminus B\), the following inequality holds:
    \begin{equation}\label{eqn:submodular_set_function}
        \rho(A \cup \{e\}) - \rho(A) \geq  \rho(B \cup \{e\}) - \rho(B), 
    \end{equation}
    for all \(A,B \subseteq V.\)
\end{definition}
Submodular set functions exhibit the property of diminishing returns, meaning that the marginal gain from adding an element to a set decreases as the set grows. 
\begin{definition}\label{def:monotoner_set_functions}
\cite{nemhauser1978maximizing_submodular_Set_functions} A set function \(\rho:2^V \rightarrow \R,\) where \(V\) is a ground set, is monotone if :
    \begin{equation}\label{eqn:monotone_set_function}
       A \subseteq B \implies \rho(A)\leq \rho(B), 
    \end{equation}
    for all \(A,B \subseteq V.\)
\end{definition}

\begin{remark}\label{rem:rho_equal_log_det}
   A set function \(\rho\) is said to be normalized if \(\rho(\emptyset) = 0.\). We employ normalized function \(\rho(\cdot) =  \log\det(\cdot),\) which is non negative, and satisfies the properties of submodularity and monotonicity defined in Definitions \ref{def:submodular_set_functions} and  \ref{def:monotoner_set_functions}.
\end{remark}

The choice of \(\rho(\cdot)\) facilitates the application of greedy algorithms, which are known to perform well with non-negative, monotone submodular functions and offer strong theoretical guarantees. 
For simplicity of notation, we represent marginal gain of adding a singleton \(\{e\}\) to set \(A\) as
\begin{equation}\label{eqn:marginal_gain}
    \rho\left(\{e\}|A\right) = \rho\left(A \cup \{e\}\right) - \rho\left(A\right)
\end{equation}

\subsection{Algorithms Description}
Having established the fundamental properties of normalized monotone submodular set functions, we are now positioned to discuss the specific algorithms utilized to efficiently solve the feature selection problem \eqref{eqn:subset_selection} in a decentralized multi-agent setting. The following sections provide a detailed description of the algorithms, including their implementation and theoretical guarantees.
\subsubsection{Algorithm \ref{alg:stochastic_greedy}}

The Stochastic-Greedy Algorithm \cite{mirzasoleiman2015stochastic_greedy} offers an efficient solution to the feature selection problem \eqref{eqn:subset_selection} accompanied by theoretical guarantees. The algorithm begins with an empty feature set and the initial information matrix \(\bar{\textbf{H}}_{\star}\) (Line 3). It iterates up to \(q\) times (Line 4), where each iteration involves randomly sampling a subset \(\mathcal{S}\) of size \(s\) from the remaining features (Line 5). The parameter \(\varepsilon\) is crucial as it determines the size of the sampled subset and affects the trade-off between the algorithm's computational efficiency and its approximation quality. The algorithm then selects the feature \(f\) that maximizes the marginal gain(Line 6) and adds it to the set while updating the information matrix accordingly (Line 7). This process continues until the feature set reaches the desired size, resulting in the final set \(\Phi_t\) and final information matrix (Line 8).

\subsubsection{Algorithm \ref{alg:approximate_greedy}}
The Approximate Greedy Algorithm provides a memory-efficient approach to the feature selection problem defined in~\eqref{eqn:subset_selection}. 
Unlike conventional methods that require storing full information matrices, this algorithm only maintains their traces, significantly reducing memory usage. 
It begins with an empty selected feature set and an initial empty score vector (Line~3). 
The algorithm iterates up to \(n\) times (Line~4); in each iteration, a feature \(f \in \Theta_t\) is randomly sampled from the remaining candidates (Line~5), and its score—quantified by the number of frames in which it is observed—is appended to the score vector. 
The evaluated feature is then removed from the candidate set (Line~7). 
Subsequently, the score vector \(\mathbf{z}\) is sorted in descending order (Line~8). 
Finally, the selected feature set \(\Phi_t\) is formed by choosing the features corresponding to the top-\(q\) entries of the sorted score vector (Line~9).


\subsection{Performance Analysis}

\begin{algorithm}[t]
\caption{Stochastic-Greedy Algorithm \cite{mirzasoleiman2015stochastic_greedy}}\label{alg:stochastic_greedy}
\begin{algorithmic}[1]
\State \textbf{Input}:  \(\bar{\textbf{H}}_{\star}\), \(\Theta_t\), {\(q\)}, \(\varepsilon.\)\\
\textbf{Output}: Set {\(\Phi_t\)} with \(|\Phi_t| = q.\) \\ 
Initialize: {\(\Phi_t \leftarrow \emptyset\)}, {\({\textbf{H}_{\star}} = \bar{\textbf{H}}_{\star}.\)}\\
\textbf{for} (\(k = 1\) to {\(q; k\leq q; k \leftarrow k+1\)}) \textbf{do}\\
\hspace*{0.5cm}$\mathcal{S} \leftarrow$~ a random subset obtained by sampling  \newline 
\hspace*{0.5cm}\(s=\frac{|\Theta_t|} {q}~\log\left(\frac{1} {\varepsilon}\right)\) ~ random elements from \(\Theta_t \setminus \Phi_t\)\\ 
\hspace*{0.5cm}\(f \leftarrow \underset{f \in \mathcal{S} }{\textup{argmax}}~\rho \left(\{f\} \mid \Phi_t\right)\)\\
\hspace{0.5cm} {\(\Phi_t \leftarrow \Phi_t \cup \{f\} ~\textup{and}~ {\textbf{H}}_{\star} \leftarrow {\textbf{H}}_{\star} + {\textbf{H}}_{\star}^{f}\)} \\
 \textbf{return} \(\Phi_t.\)
 \end{algorithmic}
\end{algorithm}

In the next section, we thoroughly evaluate the performance of the algorithm \ref{alg:stochastic_greedy} and provide rigorous analysis of algorithm \ref{alg:approximate_greedy}.
\subsubsection{Algorithm \ref{alg:stochastic_greedy}}
To demonstrate the theoretical efficacy of the Stochastic-Greedy Algorithm \ref{alg:stochastic_greedy}, we assess its approximation quality relative to the optimal solution \(\Phi_t^{*}\) in Theorem \ref{thm:stochastic_greedy_performance}.
\begin{theorem}\label{thm:stochastic_greedy_performance} \cite{mirzasoleiman2015stochastic_greedy}
    For a non-negative monotone submodular function \(\rho\) and \(s=\frac{|\Theta_t|} {q}~\log\left(\frac{1} {\varepsilon}\right),\) the Stochastic-Greedy  Algorithm \ref{alg:stochastic_greedy} achieves the following approximation guarantee:
    \begin{equation}\label{eqn:stochastic_greedy_guarantee}
        \mathbb{E}\left[\rho_e\left(\Phi_t\right)\right] \geq \left(1 - \frac{1}{e} - \varepsilon\right) \rho_e\left(\Phi_t^{*}\right),
    \end{equation}
    with only \(O\left(|\Theta_t| \log \frac{1}{\varepsilon}\right)\) function evaluations.
\end{theorem}
The theorem \ref{thm:stochastic_greedy_performance} guarantees near-optimal performance in expectation for an appropriately chosen sampling parameter \(\varepsilon,\) with computational complexity independent of cardinality constraint \(q^i.\) It underscores the algorithm's efficiency in terms of function evaluations, highlighting its practicality for large-scale feature selection.

\subsubsection{Algorithm \ref{alg:approximate_greedy}}
The greedy algorithm sequentially selects the element that maximizes the \emph{local expected information gain}, quantified as 
\[
\log\det(A + B) - \log\det(A).
\]
Although the greedy algorithm is effective and enjoys theoretical performance guarantees, it is computationally and memory intensive. Specifically, it requires storing the information matrices of all candidate elements and has a time complexity of \(O(nq)\). 

To mitigate these limitations and accelerate the selection process, we employ a tractable approximation of the log-determinant function, given by
\begin{equation}\label{eqn:approximate_log_det}
    \log\det(I + A) \leq \operatorname{tr}(A),
\end{equation}
which serves as a surrogate for \(\log\det(\cdot)\). In conjunction with the inequality
\begin{equation}\label{eqn:approx_trace}
    \operatorname{tr}(A^{-1}) \leq \frac{n}{\operatorname{tr}(A)},
\end{equation}
we use these relations to design a computationally efficient heuristic. Our objective is to minimize the uncertainty measure \(\operatorname{tr}\!\left(\mathbf{H}_{\star}(\Phi_t)^{-1}\right)\). To achieve this efficiently, we instead maximize its surrogate \(\operatorname{tr}\!\left(\mathbf{H}_{\star}(\Phi_t)\right)\), which significantly reduces computation time and memory usage. 

As established in Theorem~\ref{thm:trace_information_matrix}, the trace of the information matrix associated with a visual feature depends solely on the number of frames, making this surrogate an excellent candidate for efficient feature selection.


\begin{theorem}\label{thm:trace_information_matrix}
Let the vision measurement model noise satisfy the assumption in~\eqref{eqn:vision_noise_covaraince}. 
Then, for a given feature \( f \in \Theta_t \), the trace of its information matrix \(\mathbf{H}_{\star}^{f}\) is given by
\begin{equation}\label{eqn:trace_informaation_matrix}
    \operatorname{tr}\!\left(\mathbf{H}_{\star}^{f}\right) = 2n_f - 3,
\end{equation}
where \(n_f\) denotes the number of frames in which feature \(f\) is observed.
\end{theorem}


This Theorem states the trace of information matrix of a feature depends only on the number of frames in which it is visible.





\section{Conclusion}\label{sec:conclusion}


In this work, we presented two fast and memory-efficient algorithms for visual feature selection aimed at improving robot localization in unknown environments. The proposed methods enable robots to evaluate the utility of visual features without incurring the substantial computational and memory overhead typical of existing approaches. Our theoretical results demonstrate that a carefully chosen subset of features can preserve localization accuracy while significantly reducing processing latency, making the techniques well suited for onboard deployment in resource-constrained robotic systems. More broadly, this work highlights the importance of principled feature selection in visual perception pipelines and opens the door to future extensions, including multi-robot collaboration, adaptive horizon selection, and integration with modern learning-based visual front-ends.

\begin{algorithm}[t]
\caption{Surrogate Greedy Algorithm}\label{alg:approximate_greedy}
\begin{algorithmic}[1]
\State \textbf{Input}:  \(\Theta_t\), {\(q\),} score vector \(\bm{z} \in \R^n\)\\
\textbf{Output}: Set {\(\Phi_t\)} with \(|\Phi_t| = q\)  \\ 
Initialize: {\(\Phi_t \leftarrow \emptyset\)}.\\
\textbf{for} (\(k = 1\) to {\(q; k\leq n; k \leftarrow k+1\)}) \textbf{do}\\
\hspace*{0.5cm} Select a feature \(f \in \Theta_t\) \\ 
\hspace*{0.5cm} $\bm{z}(k) = n_f$ \\ 
\hspace*{0.5cm} {\(\Theta_t \leftarrow \Theta_t \setminus \{f\} \)} \\
\(\bm{z}_{\text{sort}}\) = sort~(\(\bm{z}\))\\
{\(\Phi_t \leftarrow \Phi_t \cup \{f: f \in \bm{z}(1:k)\} \)} \\
 \textbf{return} \(\Phi_t.\)
 \end{algorithmic}
\end{algorithm}


\printbibliography
\newpage
\appendix

\noindent \underline{\bf Proof of Theorem \ref{thm:stochastic_greedy_performance}:} The proof of this Theorem is adapted from \cite{mirzasoleiman2015stochastic_greedy}.
Given the optimal set \(\Phi_t^*\) and current selected set \(\Phi_t^k,\) and the set of randomly sampled elements \(\mathcal{S},\) the probability that there exists an element in \(\mathcal{S},\) which lies in optimal set \(\Phi_t^*\) and is not currently present in \(\Phi_t^k\) is given by 
\begin{align*}
    \mathbb{P}\left[\mathcal{S} \cap \left(\Phi_t^* \setminus \Phi_t^k\right) \neq  \emptyset\right]&= 1 - \mathbb{P}\left[\mathcal{S} \cap \left(\Phi_t^* \setminus \Phi_t^k\right) =  \emptyset\right]\\
    &= 1 - \left(1 - \frac{|\Phi_t^* \setminus \Phi_t^k|}{|\Theta_t \setminus \Phi_t^k|}\right)^s\\
    & \geq 1- \exp{\left(-s\frac{|\Phi_t^* \setminus \Phi_t^k|}{|\Theta_t \setminus \Phi_t^k|}\right)}\\
    &\geq 1- \exp{\left(-s\frac{|\Phi_t^* \setminus \Phi_t^k|}{|\Theta_t|}\right)}\\
    &\geq \left(1- e^{-\frac{sq}{|\Theta_t|}}\right)^{(\frac{|\Phi_t^* \setminus \Phi_t^k|}{q})}
\end{align*}
Substituting \(s=\frac{|\Theta_t^i|} {q^i}~\log\left(\frac{1} {\varepsilon}\right)\) leads to
\begin{align*}
      \mathbb{P}\left[\mathcal{S} \cap \left(\Phi_t^* \setminus \Phi_t^k\right) \neq  \emptyset\right] \geq (1-\varepsilon)\frac{|\Phi_t^* \setminus \Phi_t^k|}{q}
\end{align*}
 Using the above analysis, any element in \(\mathcal{S}\), which is also in \(\Phi_t^* \setminus \Phi_t^k\) has a nonzero probability. Thus, the expected marginal gain of a feature \(f \in \Phi_t^* \setminus \Phi_t^k\) is given by 
 \begin{align*}
     \mathbb{E}\left[\rho \left(\{f\} \mid \Phi_t\right)\right] &\geq  \frac{      \mathbb{P}\left[\mathcal{S} \cap \left(\Phi_t^* \setminus \Phi_t^k\right) \neq  \emptyset\right]}{|\Phi_t^* \setminus \Phi_t^k|}\sum_{f \in \Phi_t^* \setminus \Phi_t^k}^{}\rho \left(\{f\} \mid \Phi_t \right)\\
      &\geq \frac{(1-\varepsilon)}{q} \sum_{f \in \Phi_t^* \setminus \Phi_t^k}^{}\rho \left(\{f\} \mid \Phi_t \right)
 \end{align*}
The proof can be completed as follows: In iteration \(k\), given  \(\Phi_t^k\)
 \begin{align*}
     \mathbb{E}_{k}\left[\rho \left(\{f\} \mid \Phi_t^k\right)\right] 
      &\geq \frac{(1-\varepsilon)}{q} \sum_{f \in \Phi_t^* \setminus \Phi_t^k}^{}\rho \left(\{f\} \mid \Phi_t^k \right),
 \end{align*}
 where \(\mathbb{E}_{k}\left[\cdot\right] = \mathbb{E}\left[\cdot \mid \Phi_t^k\right]\) represents expectation given \(\Phi_t^k.\)
 Since \(\log\det\) is submodular, it follows that,
\begin{align*}
     \sum_{f \in \Phi_t^* \setminus \Phi_t^k}^{}\rho \left(\{f\} \mid \Phi_t^k \right) &\geq \rho(\Phi_t^* ) - \rho(\Phi_t^k ),
\end{align*}
 which leads to 
  \begin{align*}
     \mathbb{E}_{k}\left[\rho(\Phi_t^{k+1} ) - \rho(\Phi_t^k )\right] 
      &= \mathbb{E}_{k}\left[\rho \left(\{f\} \mid \Phi_t^k\right)\right] \\
      &\geq \frac{1-\varepsilon}{q} \left(\rho(\Phi_t^* ) - \rho(\Phi_t^k )\right).
 \end{align*}
 Finally, by taking expectation over \(\Phi_t^k\) leads to 
   \begin{align*}
     \mathbb{E}\left[\rho(\Phi_t^{k+1} ) - \rho(\Phi_t^k )\right] 
      &\geq \frac{1-\varepsilon}{q} \mathbb{E}\left[\rho(\Phi_t^* ) - \rho(\Phi_t^k )\right] .
 \end{align*}

 The result then follows by induction as follows:
 \begin{align*}
     \mathbb{E}\left[\rho(\Phi_t)\right] &\geq \left(1 - \left(1- \frac{1-\varepsilon}{q}\right)^q\right) \rho(\Phi_t^*)\\
     &\geq \left(1- e^{-(1-\varepsilon)}\right) \rho(\Phi_t^*)\\
     &\geq \left(1- \frac{1}{e} - \varepsilon\right) \rho(\Phi_t^*)
 \end{align*}
\hfill$\square$

\noindent \underline{\bf Proof of Theorem \ref{thm:trace_information_matrix}:}
Consider the information matrix of feature given in \eqref{eqn:Hf_mat}
\begin{equation*}\label{eqn:Hf_mat}
    \scalebox{0.85}{$
        \hspace{-0.3cm}\textup{\bf{H}}_{\star}^f = \sigma_i^{-2}\Bigg(\left(\textbf{F}_{\star}^{f}\right)^T\textbf{F}_{\star}^{f} - \left(\textbf{F}_{\star}^{f}\right)^T\textbf{E}_{\star}^{f}\left(\left(\textbf{E}_{\star}^{f}\right)^T \textbf{E}_{\star}^{f}\right)^{-1}\left(\textbf{E}_{\star}^{f}\right)^T\textbf{F}_{\star}^{f}\Bigg).
    $}
\end{equation*}
For simplicity of notation, we drop the subscripts and superscripts, and consider \(\sigma_i = 1\). 
\begin{align*}
        \text{Tr}\left(H\right)&= \text{Tr}\left((\boldsymbol{F})^T  \boldsymbol{F}\right)
                -\text{Tr}\left((\boldsymbol{F})^T \boldsymbol{E} \left((\boldsymbol{E})^T \boldsymbol{E} \right)^{-1} (\boldsymbol{E})^T \boldsymbol{F}\right)
\end{align*}
\begin{align*}
    \text{Tr}\left((\boldsymbol{F})^T  \boldsymbol{F}\right) &= \text{Tr}\left(\sum_{i =1}^{n_f}(\boldsymbol{E}_i)^T  \boldsymbol{E}_i\right)\\
    &= \sum_{i =1}^{n_f}\text{Tr}\left((\boldsymbol{E}_i)^T  \boldsymbol{E}_i\right)
    = 2n_f,
\end{align*}
where $\boldsymbol{E}_i = U_iR_i^T.$
\begin{align*}
    \text{Tr}\left((\boldsymbol{F})^T \boldsymbol{E} \left((\boldsymbol{E})^T \boldsymbol{E} \right)^{-1} (\boldsymbol{E})^T \boldsymbol{F}\right)&\\
&\hspace{-2cm}=\text{Tr}\left(\sum_{i=1}^{n_f}\left(((\boldsymbol{E}_i)^T  \boldsymbol{E}_i\right)^2\left((\boldsymbol{E})^T \boldsymbol{E} \right)^{-1} \right)\\
&\hspace{-2cm}=\text{Tr}\left(\sum_{i=1}^{n_f}\left(((\boldsymbol{E}_i)^T  \boldsymbol{E}_i\right)\left((\boldsymbol{E})^T \boldsymbol{E} \right)^{-1} \right)\\
&= 3.
\end{align*}

The result follows by combining the two pieces together. \hfill$\square$

\vspace{1mm}

\end{document}